\documentclass{article}
\usepackage{spconf,amsmath,graphicx}

 \usepackage{graphicx}
\usepackage{array}
\usepackage{color}
\usepackage{upgreek,amssymb,amsmath}
\usepackage{colortbl}
\definecolor{grey}{rgb}{0.8, 0.8, 0.84}
\definecolor{grey2}{rgb}{0.92, 0.92, 0.97}

\usepackage{subfig}
\usepackage{wrapfig}
\usepackage{hhline}

\newcolumntype{X}[1]{>{\centering\arraybackslash}p{#1}}
\newcolumntype{A}{>{\centering\arraybackslash}p{1.05cm}}
\newcolumntype{C}{>{\centering\arraybackslash}p{1.1cm}}
\newcolumntype{E}{>{\arraybackslash}p{0.7cm}}
\newcolumntype{B}{>{\centering\arraybackslash}p{1.1cm}}
\newcolumntype{Y}{>{\arraybackslash}p{1.7cm}}
\newcolumntype{D}{>{\centering\arraybackslash}p{2.75cm}}

\title{A Neural Network Approach for Mixing Language Models}
\name{Youssef Oualil, Dietrich Klakow \thanks{This research was funded by the German Research Foundation (DFG) as part of SFB 1102.}}
\address{Spoken Language Systems (LSV) \\
Collaborative Research Center on Information Density and Linguistic Encoding \\
Saarland University, Saarbr\"{u}cken, Germany \\
  {\small \tt \{firstname.lastname\}@lsv.uni-saarland.de}
}

\begin{document}

\sloppy
\ninept

\maketitle
%
\begin{abstract}
The performance of Neural Network (NN)-based language models is steadily improving 
due to the emergence of new architectures, which are able to learn
different natural language characteristics.
This paper presents a novel framework, which shows that a significant improvement can 
be achieved by combining different existing heterogeneous models in a single architecture. 
This is done through 1) a feature layer, which separately learns different NN-based models and 2) a mixture layer, 
which merges the resulting model features. In doing so, 
this architecture benefits from the learning capabilities of each model with no
noticeable increase in the number of model parameters or the training time. 
Extensive experiments conducted on the Penn Treebank (PTB) and
the Large Text Compression Benchmark (LTCB) corpus showed a significant
reduction of the perplexity when compared to state-of-the-art feedforward as
well as recurrent neural network architectures.
\end{abstract}

\vspace{-2mm}
\section{Introduction}
\label{sec:intro}

For many language technology applications such as speech recognition~\cite{Katz1987} and machine translation~\cite{Brown1990},   
a high quality Language Model (LM) is considered to be a key component to success. Traditionally, 
LMs aim to predict probable sequences of predefined linguistic units, which are typically words. 
These predictions are guided by the semantic and syntactic properties that are encoded by the LM.


The recent advances in neural network-based approaches for language modeling led to a significant improvement over the
standard $N$-gram models~\cite{Rosenfeld2000,KN1995}. This is mainly due to the continuous word representations 
they provide, which typically overcome the exponential growth of parameters that $N$-gram models require.
The NN-based LMs were first introduced by Bengio et al.~\cite{Bengio2003}, who 
proposed a Feedforward Neural Network (FNN) model as an alternative to $N$-grams. Although FNNs were shown to  
perform very well for different tasks~\cite{Schwenk2005,Goodman2001b}, their fixed context (word history) size 
constraint was a limiting factor for their performance. 
In order to overcome this constraint, 
Mikolov et al.~\cite{Mikolov2010,Mikolov2011} proposed a Recurrent Neural Network (RNN), 
which allows context information to cycle in the network. Investigating the inherent shortcomings 
of RNNs led to the Long-Short Term Memory (LSTM)-based LMs~\cite{Sundermeyer12}, which 
explicitly control the longevity of context information in the network. 
This chain of novel NN-based LMs continued with more complex and advanced models such as
Convolutional Neural Networks (CNN)~\cite{kim2016} and autoencoders~\cite{Chandar2014}, to name a few. 
%
 
LMs performance has been shown to significantly improve using model combination. 
This is typically done by either 1) designing deep networks with different architectures at the different layers, 
as it was done in~\cite{kim2016}, which combines LSTM, CNN and a highway network, or by 2) combining different models 
at the output layer, as it is done in the maximum entropy RNN model~\cite{Mikolov2011b}, which uses direct $N$-gram 
connections to the output layer, or using the classical linear interpolation~\cite{Mikolov2011c}. 
While the former category, requires a careful selection of the architectures to combine for a well-suited feature design,
and can be difficult/slow to train, the second category knows a significant increase in the number of parameters
when combining multiple models. 

Motivated by the work in~\cite{Mikolov2011b}, we have recently proposed a Sequential Recurrent Neural Network (SRNN)~\cite{oualil2016},
which combines FFN information and RNN. 
In this paper, we continue along this line of work by proposing a generalized 
framework to combine different heterogeneous NN-based architectures in a single mixture model.
More particularly, the proposed architecture uses 1) a hidden feature layer to, separately, learn each of the models 
to be combined, and 2) a hidden mixture layer, which combines the resulting model features.
Moreover, this architecture uses a single word embedding matrix, which 
is learned from all models, and a single output layer. This framework is, in principle, able to 
combine different NN-based LMs (e.g., FNN, RNN, LSTM, etc.) with no direct constraints 
on the number of models to combine or their configurations.

We proceed as follows. Section~\ref{sec:NN} presents an overview of the basic NN-based LMs. 
Section~\ref{sec:NNMM} introduces the proposed neural mixture model. 
Then, Section~\ref{sec:EXP} evaluates the proposed network in comparison to different 
state-of-the-art language models for perplexity on the PTB and the LTCB corpus. Finally, 
we conclude in Section~\ref{sec:CC}.

\section{Neural Network Language Models}
\label{sec:NN}
The goal of a language model is to estimate the probability distribution $p(w_1^T)$ of word sequences $w_1^T = w_1,\cdots,w_T$.
Using the chain rule, this distribution can be expressed as
\vspace{-1mm}
\begin{equation}
  \label{eq:prob}
\displaystyle{ p(w_1^T) = \prod_{t=1}^T{p(w_t|w_1^{t-1})} }
\vspace{-2mm}
\end{equation}
Let $U$ be a word embedding matrix and let $W$ be the hidden-to-output weights. 
NN-based LMs (NNLMs), that consider word embeddings as input, approximate each of the 
terms involved in this product in a bottom-up evaluation of the network according to
  \vspace{-1mm}
\begin{align}
\label{eqn:eqm-1}
 H^t &= \mathcal{M}(\mathcal{P}, \mathcal{R}^{t-1}, U) \\  
\label{eqn:eqm-2}
 O^t  &= g \left( H^{t} \cdot W\right)
  \vspace{-1mm}
 \end{align}
where $\mathcal{M}$ represents a particular NN-based model, which can be a deep architecture, 
$\mathcal{P}$ denotes its parameters and $\mathcal{R}^{t-1}$ denotes its recurrent information at time $t$.
$g(\cdot)$ is the softmax function.

The rest of this section briefly introduces $\mathcal{M}$, $\mathcal{P}$ and $\mathcal{R}^{t-1}$ 
for the basic architectures, namely FNN, RNN and LSTM, which were investigated and evaluated as 
different components in the proposed mixture model. The proposed architecture, however, is general 
and can include all NNLMs that consider world embeddings as input.

\subsection{Feedforward Neural Networks}
\label{ssec:FNN}
Similarly to $N$-gram models, FNN uses the Markov assumption of order $N-1$
to approximate~(\ref{eq:prob}). That is, the current word
depends only on the last $N-1$ words. Subsequently, $\mathcal{M}$ is given by 
\vspace{-0.5mm}
\begin{align}
\label{eqn:eqfnn-1}
 E^{t-i} &=  X^{t-i} \cdot U \hspace{1mm}, \quad \quad i=N-1,\cdots,1  \\  
\label{eqn:eqfnn-2}
 H^{t}   &= f \left( \sum_{i=1}^{N-1} E^{t-i} \cdot V^i \right) 
 \end{align}
$X^{t-i}$ is a one-hot encoding of the word $w_{t-i}$.
Thus, $E^{t-i}$ is the continuous representation of the word $w_{t-i}$.
$f(\cdot)$ is an activation function.
Hence, for an FNN model $\mathcal{M}$, $\displaystyle{\mathcal{P}=\{V^i\}_{i=1}^{N-1}}$ 
and $\displaystyle{\mathcal{R}^{t-1}=\emptyset}$.

\subsection{Recurrent Neural Networks}
\label{ssec:RNN}
RNN attempts to capture the complete history in a context vector 
$h_{t}$, which represents the state of the network and evolves in time.
Therefore, RNN approximates each term in~(\ref{eq:prob}) as
$\displaystyle{p(w_t|w_{1}^T) \approx p(w_t|h_{t})}$. As a result,
$\mathcal{M}$ for an RNN is given by 
%
\begin{equation}
  \label{eq:eqrnn-1}
	  H^{t} = f \left( X^{t-1} \cdot U + H^{t-1} \cdot V \right)
\end{equation}
Thus, for an RNN model $\mathcal{M}$, $\displaystyle{\mathcal{P}=V}$ 
and $\displaystyle{\mathcal{R}^{t-1}=H^{t-1}}$.

\subsection{Long-Short Term Memory Networks}
\label{ssec:LSTM}

In order to alleviate the rapidly changing context issue in standard RNNs and control 
the longevity of the dependencies modeling in the network, the LSTM architecture~\cite{Sundermeyer12}
introduces an internal memory state $C^{t}$, which explicitly controls the amount of information, 
to forget or to add to the network, before estimating the current hidden state. 
Formally, an LSTM model $\mathcal{M}$ is given by 
\vspace{-1mm}
\begin{align}
  \label{eqn:eqlstm-1}
   E^{t-1} &= X^{t-1} \cdot U \\
  \label{eqn:eqlstm-2}
  {\lbrace i,f,o\rbrace}^t &= \sigma \left(  V_w^{i,f,o} \cdot E^{t-1} + V_h^{i,f,o} \cdot H^{t-1} \right) \\
  \label{eqn:eqlstm-3}
  \tilde{C^{t}} &= f \left( V_w^c \cdot E^{t-1} +  V_h^c \cdot H^{t-1} \right) \\
  \label{eqn:eqlstm-4}
  C^t &= f^t \odot C^{t-1} + i^t \odot \tilde{C^{t}}  \\
  \label{eqn:eqlstm-5}
  H^t &= o^t \odot f \left(C^t \right) 
 \end{align}
where $\odot$ is the element-wise product, $\tilde{C^{t}}$ is the memory candidate, 
whereas $i^t,f^t$ and $o^t$ are the input, forget and output gates of the network, respectively. 
Hence, for an LSTM model $\mathcal{M}$, $\displaystyle{\mathcal{R}^t=\{H^t,C^t\}}$ 
and $\displaystyle{\mathcal{P}=\{V_w^{i,f,o,c},V_h^{i,f,o,c}\}}$.

%
\section{Neural Network Mixture Models}
\label{sec:NNMM}

On the contrary to a large number of research directions on improving or designing 
(new) particular neural architectures for language modeling, the work presented in this paper is 
an attempt to design a general architecture, which is able to combine different types of existing 
heterogeneous models rather than investigating new ones.  

\subsection{Model Combination for Language Modeling}
\label{ssec:MC}

The work presented in this paper is motivated by recent research showing that model combination can lead to a significant improvement 
in LM performance~\cite{Mikolov2011c}. This is typically done by either 1) designing deep networks with different architectures at the different layers, 
as it was done in~\cite{kim2016}. This category of model combination, however, requires a careful selection of the architectures 
to combine for a well-suited feature design, as it can be difficult/slow to train, whereas the second category 2) combines different models 
at the output layer, as it is done in the maximum entropy RNN model~\cite{Mikolov2011b} or using the classical linear 
interpolation~\cite{Mikolov2011c}. This category typically leads to a significant increase in the number of parameters 
when combining multiple models.
 
In a first attempt to circumvent these problems, we have recently proposed an SRNN model~\cite{oualil2016},
which combines FFN information and RNN through additional sequential connections at the hidden layer. Although SRNN
was successful and did not noticeably suffer from the aforementioned problems, it was solely designed to combine RNN and 
FNN and is, therefore, not well-suited for other architectures. This paper continues along this line of work by proposing
a general architecture to combine different heterogeneous neural models with no direct constraints on the number or type of models.

\subsection{Neural Network Mixture Models}
\label{ssec:MNN}

This section introduces the mathematical formulation of the proposed mixture model.
Let $\displaystyle{\{\mathcal{M}_m\}_{m=1}^{M}}$ be a set of $M$ models to combine, and let
$\displaystyle{\{\mathcal{P}_m, \mathcal{R}_m^t\}_{m=1}^{M}}$ be their corresponding model 
parameters and recurrent information at time $t$, respectively. For the basic NNLMs, namely 
FNN, RNN and LSTM, $\mathcal{M}_m$, $\mathcal{P}_m$ and $\mathcal{R}_m^t$ were introduced in 
Section~\ref{sec:NN}.
%
%
%

\begin{figure}[!h]
\vspace{-1mm}
  \centering
  \includegraphics[trim=00 00 00 0 ,clip, totalheight=0.21\textheight, width=0.9\linewidth]{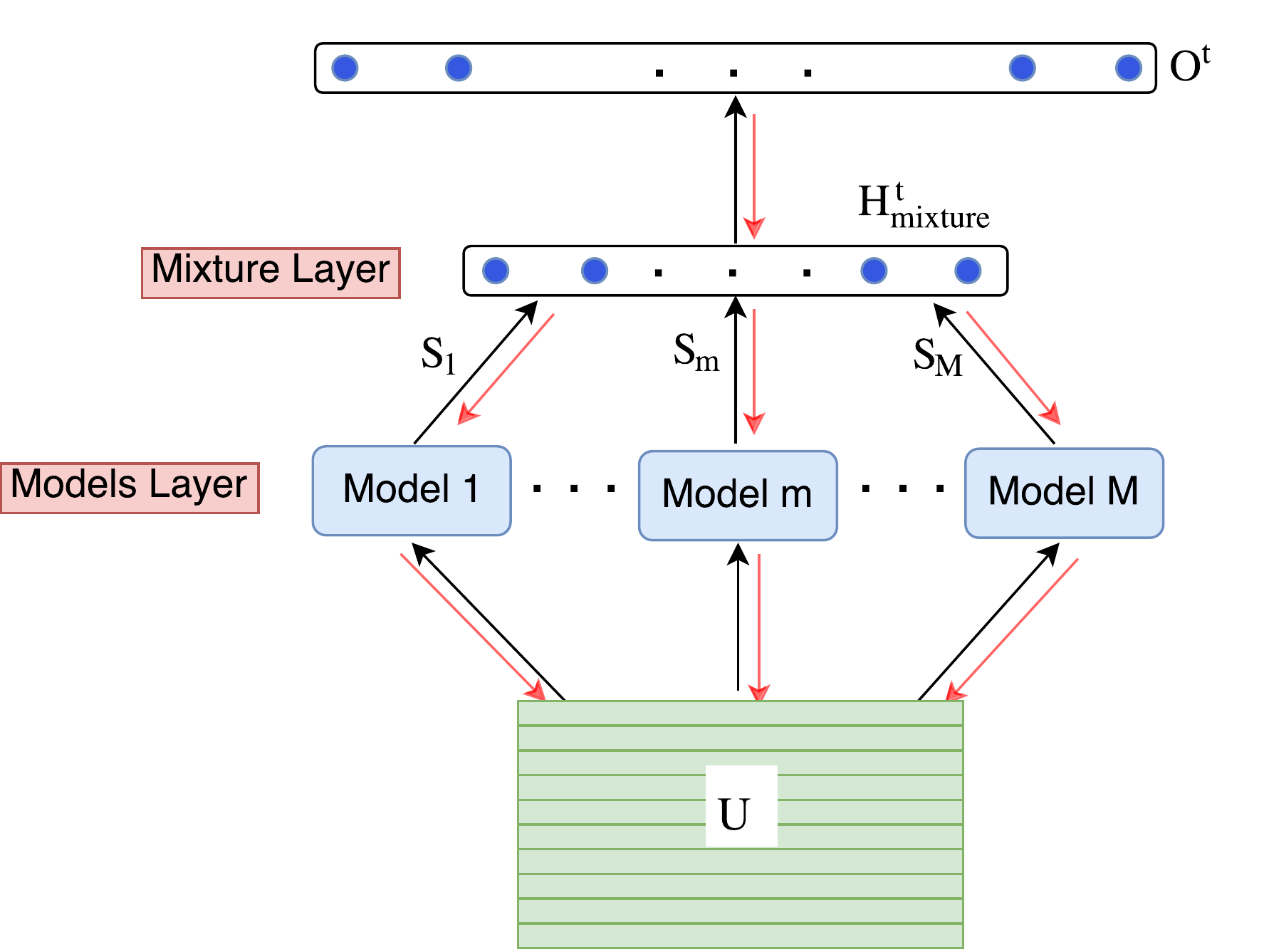}
   \vspace{-0.8mm}
	\caption{Neural Mixture Model (NMM) architecture. Red (back) arrows show the error propagation during training.}
  \label{fig:NMM}
\vspace{-2mm}
\end{figure}

Let $U$ be the shared word embedding matrix, which is learned 
during training from all models in the mixture. The mixture model is given 
by the following steps (see illustration in Fig.~\ref{fig:NMM}):

\noindent \textit{1) Feature layer: update each model and calculate its features} 
\begin{equation}
  \label{eqn:label}
    H_m^t = \mathcal{M}_m(\mathcal{P}_m, \mathcal{R}_m^{t-1}, U), \quad m=1,\cdots,M
\end{equation}
\noindent \textit{2) Mixture layer: combine the different features}
\begin{equation}
    H_{mixture}^t = f_{mixture} \left( \sum_{m=1}^{M} H_m^t \cdot S_m \right) \label{mix:mixture}
\end{equation}
\noindent \textit{3) Output layer: calculate the output using a softmax function}
\begin{equation}
    O^t  = g \left( H_{mixture}^t \cdot W\right) 
\end{equation}

$f_{mixture}$ is a non-linear mixing function, whereas $S_m,$ $m=1,\cdots,M$ are the mixture weights (matrices).

Although the experiments conducted in this work mainly include FNN, RNN and LSTM, the set of possible model selection for $\mathcal{M}_m$  
is not restricted to these but includes all NN-based models that take word embeddings as input. 

The proposed mixture model uses a single word embedding matrix and a single output layer with predefined and fixed sizes. 
The latter are independent of the sizes of the mixture models. In doing so, this model does not suffer from the significant 
parameter growth when increasing the number of models in the mixture. We can also see that this architecture does not 
impose any direct constraints on the number of models to combine, their size or their type. Hence, we can combine, 
for instance, models of the same type but with different sizes/configurations, as we can combine heterogeneous models 
such as recurrent and non-recurrent models, in a single mixture. Moreover, the mixture models can also be deep 
architectures with multiple hidden layers. 

\vspace{-2mm}
%
%
\subsection{Training of Neural Mixture Models}
\label{ssec:Training}

NMM training follows the standard back-propagation algorithm used to train neural architectures. 
More particularly, the error at the output layer is propagated to all models in the mixture. 
At this stage, each model receives a network error, updates its parameters, and propagates its error to the 
shared word embedding (input) layer. We should also mention here that recurrent models can be ``unfolded'' in time, 
independently of the other models in the mixture, as it is done for standard networks.   
Once each model is updated, the continuous word representations are then updated as well while taking into
account the individual network errors emerging from the different models in the 
mixture (see illustration in Fig.~\ref{fig:NMM}).  

The joint training of the mixture models is expected to lead to a ``complementarity'' effect. 
We mean by ``complementarity'' that the mixture models perform poorly when evaluated separately but lead to a much better performance
when tested jointly. This is typically a result of the models learning and modeling, eventually, different features.
Moreover, the joint learning is also expected to lead to a richer and more expressive word embeddings.
 
\subsection{Model Dropout}
\label{ssec:MD}

In order to 1) enforce models co-training and 2) avoid network over-fitting
when the number of models in the mixture is large. We
use a model dropout technique, which is inspired by the standard 
dropout regularization~\cite{Srivastava14} that is widely used to train neural networks. The idea here is to 
have ``models'' replace ``neurons'' in the standard dropout. Therefore, for each training example,
a model is to be dropped with a probability $p_d$. Then, only models that are selected contribute to the mixture
and have their parameters and mixing weights $S_m$ updated. Similarly to standard dropout, 
model dropout is applied only to non-recurrent models in the mixture.

    \section{Experiments and Results}
    \label{sec:EXP}
    \subsection{Experimental Setup}
    We evaluated the proposed architecture on two different benchmark tasks. 
    The first set of experiments was conducted on the Penn Treebank (PTB) corpus using the standard division, 
    e.g.~\cite{Mikolov2011,FOFE2015}; sections 0-20 are used for training
    while sections 21-22 and 23-24 are used for validation and testing. The vocabulary was limited 
    to the 10k most frequent words while the remaining
    words were all mapped to the token $<$unk$>$. In order to evaluate how the proposed
    approach scales to large corpora, we run a set of experiments on
    the Large Text Compression Benchmark (LTCB)~\cite{Mahoney2011}.
    This corpus is based on the enwik9 dataset which contains the first $10^9$ 
    bytes of enwiki-20060303-pages-articles.xml. We adopted the same 
    training-test-validation data split and pre-processing from~\cite{FOFE2015}. 
    The vocabulary was limited to the 80k most frequent words.
    Details about the sizes of these two corpora and the percentage of 
    Out-Of-Vocabulary (OOV) words that were mapped to $<$unk$>$ can be found in Table~\ref{tab:corpora}.
    \vspace{-1mm}
    \renewcommand{\tabcolsep}{2.2pt}
    \begin{table}[!h]
    \caption{\label{tab:corpora} {\it Corpus size in number of words and $<$unk$>$ rate.}}
    \centerline{
      \begin{tabular}{| c || c | c || c | c || c | c | }
      \hline
      \multicolumn{1}{|c||}{} & \multicolumn{2}{c||}{Train} & \multicolumn{2}{c||}{Dev} & \multicolumn{2}{c|}{Test} \\
	\hline
      Corpus & \#W & $<$unk$>$ & \#W & $<$unk$>$ & \#W & $<$unk$>$ \\
	\hline
      \hline
      PTB  & 930K & 6.52\% & 82K & 6.47\% & 74K & 7.45\% \\
	\hline
      LTCB & 133M & 1.43\% & 7.8M & 2.15\% & 7.9M & 2.30\%  \\
	\hline
      \end{tabular}}
    \end{table}
    %
		
    The results reported below compare the proposed Neural Mixture Model (NMM) 
    approach to the baseline NNLMs.
    In particular, we compare our model to the FNN-based LM~\cite{Bengio2003}, 
    the full RNN~\cite{Mikolov2011} (without classes) as well as RNN with maximum entropy (RNNME)~\cite{Mikolov2011b}.
    We also report results for the LSTM architecture~\cite{Sundermeyer12}, 
		and the recently proposed SRNN model~\cite{oualil2016}.
    
    Although the proposed approach was not designed for a particular mixture of 
    models, we only report results for different combinations of FNN, RNN and LSTM, 
    which are considered to be the baseline NNLMs. 
    For clarity, an NMM result is presented as  
    $\displaystyle{F_{S_1,\cdots,S_f}^{N_1,\cdots,N_f}+R_{S_1,\cdots,S_r}^{ }+L_{S_1,\cdots,S_l}^{ }}$, 
    where $f$ is the number of FNNs in the mixture, $S_m, m=1,\cdots,f$ are their corresponding 
    hidden layer sizes (that are fed to the mixture) and $N_m, m=1,\cdots,f$ are their fixed history sizes. 
    The same notation holds for RNN and LSTM, where $r$ and $l$ are the number of RNNs and LSTMs in 
    the mixture, respectively, and $N_r,N_l=1$. The number of models in the mixture is given by $f+r+l$.
    Moreover, the notation $\displaystyle{F_{S_f}^{N_b-N_e}}$ means that this model combines $N_e-N_b+1$ consecutive FNN models 
    with respective history sizes $N_b, N_b+1, \cdots,N_e$, with all models having the same hidden layer size $S_f$.
    
    \subsection{PTB Experiments}
    
    For the PTB experiments, all models have a hidden layer size of 400,
    with FFNN and SRNN using the Rectified Linear Unit (ReLu) i.e., $f(x)=max(0,x)$ 
    as activation function and having 2 hidden layers. ReLu is also used as activation function
		for the mixture layer in NMMs, which use a single hidden layer.
    The embedding size is 100 for SRNN and NMMs, whereas it is set to 400 for RNN and 200 for FNN and LSTM. 
    The training is performed using the stochastic gradient descent algorithm with a mini-batch size of 200.
    the learning rate is initialized to 0.4, the momentum is set to 0.9, the weight 
    decay is fixed at $4$x$10^{-5}$, the model dropout is set to $0.4$ and the training is done in epochs. 
    The weights initialization follows the normalized initialization proposed in~\cite{Glorot2010}. 
    Similarly to~\cite{Mikolov2010}, the learning rate is halved when no significant improvement 
    in the log-likelihood of the validation data is observed. 
  The BPTT was set to 5 time steps for all recurrent models. 
  In the tables below, the results  are reported
  in terms of perplexity (PPL), Number of model Parameters (NoP) and 
  the Parameter Growth (PG) for NMM, which is defined as the relative increase in the 
  number of parameters of NMM w.r.t. the baseline model in the table.  
	In order to demonstrate the power of the joint training, we also report the perplexity PPL 
	and NoP of the Linearly Interpolated (LI) models in the mixture after training them separately. 
	In this case, each model learns its own word embedding and output layer.  

  %
  \renewcommand{\tabcolsep}{2.4pt}
  \begin{table}[!h]
  \caption{\label{tab:ptb} {\it LMs performance on the PTB test set.}}
  \vspace{-2mm}
  \centerline{
		\begin{tabular}{| Y | A | A | C || A | A |}
    \hline
    model & PPL &  NoP    & PG  & PPL(LI) & NoP(LI) \\
      \hline
       FNN (N=5)                   & 114 & 6.49M  &    ---    &   ---   &  ---   \\
      \hline
    $\displaystyle{F^{2,3}_{200}}$ & 117 & 5.27M  & -18.80\%  &  120.0  & 6.10M  \\ 
      \hline
    $\displaystyle{F^{2-5}_{200}}$ & 110 & 5.61M  & -13.56\%  &  112.0  & 12.28M  \\ 
      \hline
    \hline
    LSTM                                  &     105      &  6.97M  &   ---   &   ---  &   ---    \\
      \hline
    $\displaystyle{L_{100}+F^{2}_{200}}$  & \textbf{102} &  5.25M  & -24.68\% &  114  &  5.12M   \\ 
      \hline
    $\displaystyle{L_{100}+R_{100}}$      & \textbf{102} &  5.18M  & -25.68\% &  118  &  4.09M  \\ 
      \hline
    \hline
    RNN                                    & 117  &  8.17M  &   ---    &  ---  &  --- \\
      \hline
    $\displaystyle{R_{100}+F^{2}_{200}}$   & 109  &  5.18M  & -36.60\% &  119  & 5.05M  \\ 
      \hline
    $\displaystyle{R_{100}+F^{2-6}_{200}}$ & 105  &  5.86M  & -28.27\% &  108  & 17.41M \\ 
      \hline
     \hline
     RNNME     & 117   &   10G   & --- & --- &  --- \\
     \hline
     WD-SRNN   & 104   &  6.33M  & --- & --- &  --- \\ 
    \hline
    WI-SRNN    & 104   &  5.33M  & --- & --- &  --- \\ 
      \hline
    \end{tabular} }
    \vspace{-1mm}
  \end{table}
  The PTB results reported in Table~\ref{tab:ptb} show clearly that combining different small-size models with a reduced word embedding size
	results in a better perplexity performance compared to the baseline models, with a significant decrease in the NoP 
	required by the mixture. More particularly, we can see that adding a single FNN model to a small size LSTM or RNN
	is sufficient to outperform the baseline models while reducing the number of parameters by $24\%$ and $36\%$, respectively. 
	The same conclusion can be drawn when combining an RNN with an LSTM.  
	We can also see that adding more FNN models to each of these mixtures leads to additional improvements while keeping
	the number of parameters significantly small. Table~\ref{tab:ptb} also shows that training the small size models (in the mixture)
	separately, and then linearly interpolating them, results in a slightly worse performance compared to the mixture model with a noticeable 
	increase in the NoP. This conclusion emphasizes the importance of the joint training. Moreover, we can also see that mixing RNN and FNNs
	leads to a comparable performance to SRNN, which was particularly designed to enhance RNN with FNN information. 
	The proposed approach, however, does not particularity encode the individual characteristics of the models in the mixture, 
	which reflects its ability to include different types of NNLMs. We can also conclude that combining FNN with recurrent models 
	leads to a more significant improvement when compared to mixtures of FNNs. This conclusion shows, similarly to other work 
	e.g.~\cite{oualil2016,Mikolov2011b}, that recurrent models can be further improved using N-gram/feedforward information, 
	given that they model different linguistic features.
  \begin{figure}[!ht]
  \vspace{-1mm}
  \centering
  \includegraphics[trim=36 0 74 10 ,clip, totalheight=0.2\textheight, width=1.0\linewidth]{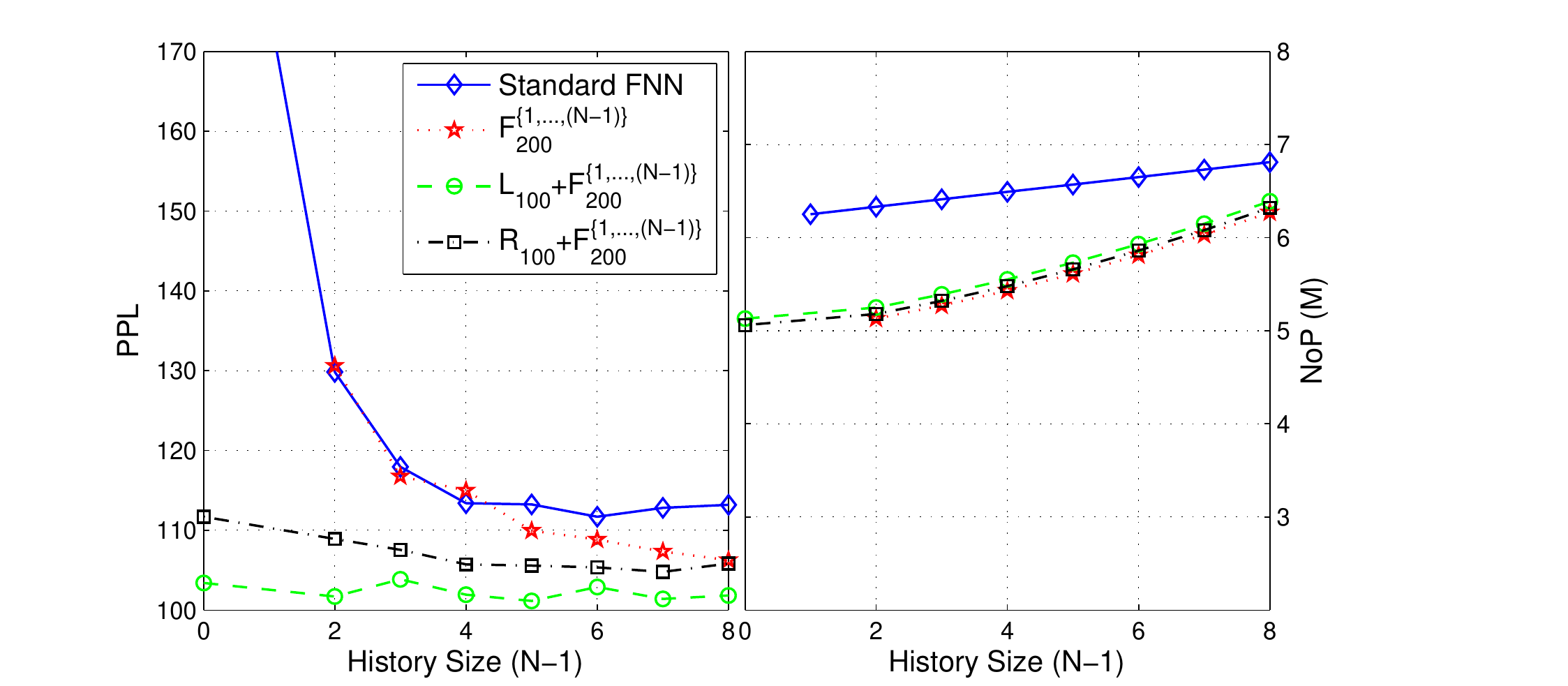}
  \vspace{-3mm}
  \caption{Perplexity vs parameter growth of different mixture models while iteratively adding more FNN models to the mixture.}
  \label{fig:NMMR}
  \vspace{-3mm}
  \end{figure}

  Fig.~\ref{fig:NMMR} is an extension of Table~\ref{tab:ptb}, which shows the change in the perplexity and NoP of different NMMs when iteratively adding
  more FFN models to the mixture. This figure confirms that combining heterogeneous models (combining LSTM or RNN with FNNs) achieves a better performance compared 
  to combining only FNN models. We can also conclude from this figure that the improvement 
	becomes very slow after adding 4 FNN models to each mixture.
  \subsection{LTCB Experiments}
  
  The LTCB experiments use the same PTB setup with minor changes. 
  The results shown in Table~\ref{tab:ltcb} follow the same experimental setup used in~\cite{oualil2016}. More precisely, 
  these results were obtained without usage of momentum, model dropout or weight decay whereas the mini-batch size was set to 400.
  The FNN architecture contains 2 hidden layers of size 600 whereas RNN, LSTM, SRNN and NMM have 
  a single hidden layer of size 600. 

  \vspace{-1mm}
  \renewcommand{\tabcolsep}{2.7pt}
  \begin{table}[!h]
  \caption{\label{tab:ltcb} {\it LMs Perplexity on the LTCB test set.}}
  \vspace{-2mm}
  \centerline{
    \begin{tabular}{| c | B | B | B | }
    \hline
    model & PPL &  NoP    & PG  \\
    \hline
    FNN[4*200]-600-600-80k         & 110 & 64.92M  & --- \\
     \hline
    $\displaystyle{F^{2-4}_{600}}$ & 102 & 66.24M  & 2.03\% \\ 
      \hline
    $\displaystyle{F^{2-7}_{100}}$ & 92  & 64.98M  & 0.09\% \\ 
      \hline
     \hline
      RNN[600]-600-80k                     &  85  &  96.44M  & --- \\
     \hline
    $\displaystyle{R_{200}+F^{4}_{400}}$   &  84  &  64.80M  & -32.81\% \\ 
      \hline
    $\displaystyle{R_{200}+F^{2-4}_{600}}$ &  77  &  66.40M  & -31.15\% \\ 
      \hline
      \hline
    LSTM[600]-600-80k                    &  66 &  66.00M  & --- \\
      \hline
		$\displaystyle{L_{400}+R_{200}}$    &  64 &  65.44M  & -1.51\% \\ 
      \hline
    $\displaystyle{L_{300}+F^{4}_{300}}$ &  64 &  65.28M  & -1.75\% \\ 
      \hline
    $\displaystyle{L_{600}+F^{4}_{600}}$ &  \textbf{58} &  67.21M  & 1.16\% \\ 
      \hline
      \hline
	  WI-SRNN[4*200]-600-80k    &  77   & 64.56M  & --- \\
      \hline
    WD-SRNN[4*200]-600-80k    &  72   & 80.56M  & --- \\
      \hline
    \end{tabular} }
  \end{table}
  \vspace{-1mm}
  The LTCB results shown in Table~\ref{tab:ltcb} generally confirm 
  the PTB conclusions. In particular, we can see that combining recurrent models, with half (third for RNN) their 
	original size, with a single FNN model leads to a comparable performance to the baseline models. 
	Moreover, increasing the mixture models size (for LSTM) or increasing the number of FNNs (for RNN) improves the 
	performance further with no noticeable increase in the NoP. 
  Similarly to the PTB, we can also see that NMM achieves the same performance as the 
	WI-SRNN model with a NoP reduction of $31\%$ compared to the original RNN model. 
	The FFN mixture results show a more significant improvement when combining multiple small-size (100)
	models compared to mixing few large models (600). This conclusion shows that the strength of mixture models
	lies in their ability to combine the learning capabilities of different models, even with small sizes.

\section{Conclusion and Future Work}
\label{sec:CC}

We have presented a neural mixture model which
is able to combine heterogeneous NN-based LMs in a single architecture.
Experiments on PTB and LTCB corpora have shown that this architecture substantially 
outperforms many state-of-the-art neural systems, due to its ability to combine 
learning capabilities of different architectures.
Further gains could be made using a more advanced model selection or feature 
combination at the mixing layer instead of the simple model weighting. 
These will be investigated in future work.

\vfill\pagebreak

\bibliographystyle{IEEEbib}
\bibliography{strings,refs}

\end{document}